\documentclass[letterpaper]{article} 
\usepackage[preprint]{aaai2027}  
\usepackage[hyphens]{url}  
\usepackage{graphicx} 
\usepackage{natbib}  
\usepackage{caption} 
\usepackage{algorithm}

\usepackage{newfloat}
\usepackage{listings}
\DeclareCaptionStyle{ruled}{labelfont=normalfont,labelsep=colon,strut=off} 
\floatstyle{ruled}
\newfloat{listing}{tb}{lst}{}
\floatname{listing}{Listing}

\usepackage{booktabs}
\usepackage{algpseudocode}
\usepackage{amssymb}
\usepackage{amsmath}
\usepackage{adjustbox}
\usepackage[table]{xcolor}
\usepackage{makecell}
\usepackage{enumitem}
\usepackage{multirow}
\usepackage{pifont}
\usepackage{mathrsfs}
\usepackage{xr}

\newif\ifshowrevisions

\title{PEARL: Training Socratic Tutors with Pedagogically Aligned Reinforcement Learning}
\author{
    Qikai Chang\textsuperscript{\rm 1},
    Zhenrong Zhang\textsuperscript{\rm 1,\rm 2},
    Linbo Chen\textsuperscript{\rm 1},
    Pengfei Hu\textsuperscript{\rm 1},\\
    Jianshu Zhang\textsuperscript{\rm 2},
    Youhui Guo\textsuperscript{\rm 2},
    Jun Du\textsuperscript{\rm 1}\corresponding
}
\affiliations{
    \textsuperscript{\rm 1}University of Science and Technology of China, 
    \textsuperscript{\rm 2}iFLYTEK Research\\
    jundu@ustc.edu.cn
}

\begin{document}

\maketitle

\begin{abstract}
Large Language Models (LLMs) show strong potential as educational tutors. Existing approaches typically train them to solve problems and provide correct answers, but this problem-solving-centered paradigm overlooks key requirements of effective tutoring: progressive guidance and the coordination of multiple pedagogical objectives across multi-turn interactions. Developing such tutors remains challenging because student behavior varies substantially with individual knowledge states, pedagogical effectiveness depends on multiple factors beyond final-answer correctness, and coordinating these objectives over tutor-student interactions is inherently difficult. To address these challenges, we propose PEARL, a \textbf{PE}dagogically \textbf{A}ligned \textbf{R}einforcement \textbf{L}earning framework for training Socratic tutoring agents. First, we introduce a controllable student simulator that disentangles latent cognitive states from response generation, enabling simulation of diverse abilities and misconceptions. Second, we develop a pedagogically aligned reward model that jointly assesses pedagogical quality and objective correctness. Finally, we propose a stable multi-objective reinforcement learning approach that balances competing pedagogical objectives during tutor training. Experiments across multiple benchmarks show that PEARL performs competitively against tutoring-specific open-source systems and leading proprietary LLMs.
\end{abstract}


\begin{figure}[t]
	\centering
	\includegraphics[width=0.95\linewidth]{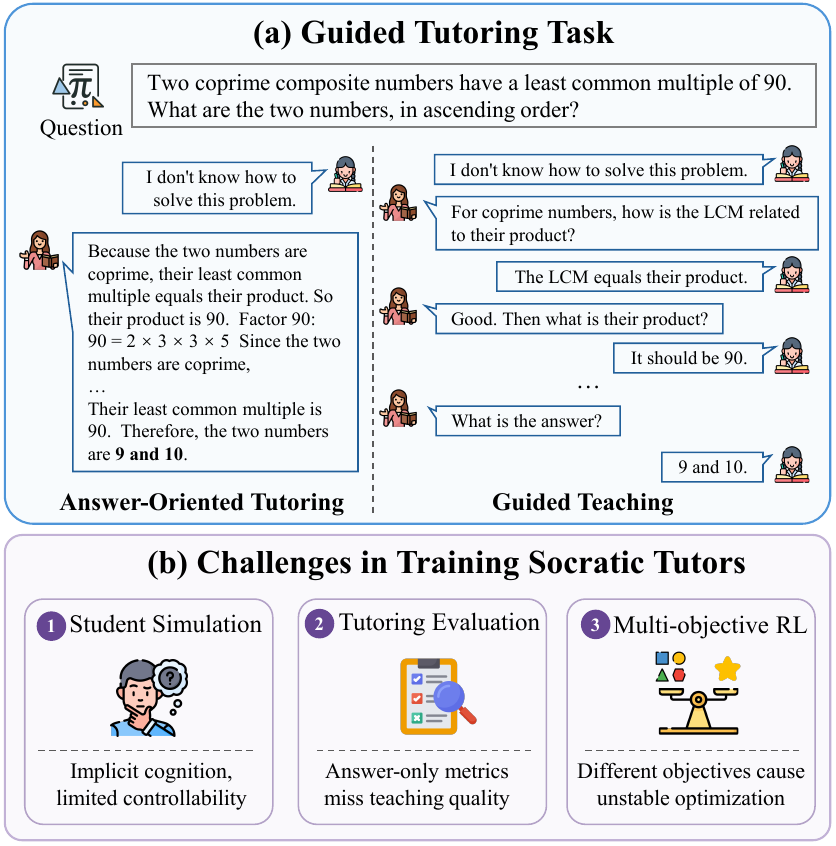}
	\caption{Overview of the guided tutoring task and its challenges. (a) Guided teaching elicits student reasoning through Socratic interaction. (b) Training such tutors requires controllable simulation, holistic evaluation, and stable multi-objective RL.}
	\label{fig:multi_agent_interaction}
\end{figure}

\section{Introduction}


In recent years, Large Language Models (LLMs) have achieved remarkable progress in reasoning, self-reflection, and generalization \citep{yang2025qwen3, bai2025qwen3}. With capabilities such as multi-turn dialogue, planning, memory augmentation, and tool invocation, LLM-based agents have shown strong potential across numerous domains \citep{team2026kimi, zeng2026glm, liu2025deepseek, chang2026thor}. Among these applications, education is especially promising because it relies on interactive dialogue and reasoning-intensive problem solving \citep{letourneau2025systematic}.

Educational agents aim to support students through one-on-one, multi-turn dialogue guided by pedagogical principles \citep{vajjala2025opportunities, park2024empowering}, as shown in Figure~\ref{fig:multi_agent_interaction}(a). However, strong problem-solving ability does not ensure effective teaching. Rather than simply providing answers, effective tutors engage students in guided, multi-turn dialogue to diagnose their misconceptions and provide heuristic guidance, following a Socratic approach to instruction \citep{hu2025generative, bonino2024euler, zhang2024spl}. Developing such pedagogical intuition remains a central challenge for LLM-based tutoring systems.

Although recent studies have made progress in developing LLM-based tutors \citep{park2024empowering, zhang2024spl, shi2025educationq, liu2025one, wang2025llm}, training tutoring agents still faces three core bottlenecks, as illustrated in Figure~\ref{fig:multi_agent_interaction}(b): \textbf{weakly controllable student simulation, under-specified pedagogical reward modeling, and unstable multi-objective optimization}. (1) For \textbf{student simulation}, existing systems prompt LLMs to role-play students \citep{macina2023mathdial, liu2024socraticlm, dinucu2025problem, sonkar2024llm, xu2025classroom}, sometimes augmented with student profiles \citep{liu2024personality, gao2025agent4edu}. End-to-end role-play makes it difficult to control knowledge mastery separately from behavioral tendencies. It can also yield narrow or simulator-specific interaction patterns. (2) For \textbf{pedagogical reward modeling}, existing evaluation methods either rely on costly human annotation \citep{macina2023mathdial} or adopt LLM-as-a-judge frameworks \citep{wei2025elmes, song2025cultivating, zheng2023judging, gu2024survey, maurya2025unifying}. However, these evaluation signals are not always well aligned with educational principles, as they tend to emphasize correctness while failing to jointly assess the quality of the pedagogical process and the provision of affective support. (3) For \textbf{multi-objective optimization}, tutoring involves potentially competing objectives, including correctness, answer-leakage control, completeness, and adaptivity. Directly averaging terminal rewards can allow objectives with high-variance or dense reward signals to dominate policy updates in methods such as GRPO \citep{shao2024deepseekmath} and PPO \citep{schulman2017proximal}.

To address these challenges, we propose PEARL, a \textbf{PE}dagogically \textbf{A}ligned \textbf{R}einforcement \textbf{L}earning framework for training tutoring agents. PEARL coordinates three components. (1) A \textbf{controllable student simulator} explicitly tracks and updates students’ knowledge mastery in a prerequisite-aware network, while separating this structured knowledge state and a five-dimensional cognitive profile from behavioral-intent prediction and response realization, enabling controlled and behaviorally varied interaction trajectories. (2) A \textbf{pedagogically grounded multidimensional reward model} evaluates complete dialogues across correctness, pedagogical process, and affective support, while accounting for the student's evolving state. (3) A \textbf{multi-objective RL scheme} aggregates and discretizes rewards within each dimension before normalizing and combining advantages across dimensions, thereby reducing domination by objectives with larger variance or denser signals.

We evaluate PEARL on GSM8K \citep{cobbe2021training}, MATH-500 \citep{hendrycks2021measuring}, MathTutorBench \citep{macina2025mathtutorbench}, and MathDial \citep{macina2023mathdial} using a training-aligned reward model, an independent external evaluator, and blind teacher comparisons. PEARL substantially improves upon its backbone and outperforms evaluated tutoring-specific open-source systems, achieving the strongest performance among the evaluated open-source models and narrowing the gap to proprietary LLMs.

Our primary contributions are as follows: 
\begin{itemize}	
	\item We propose a controllable student simulator that explicitly models students' knowledge mastery, cognitive tendencies, and learning abilities, enabling controlled simulation of diverse learning behaviors.
	
	\item We develop a pedagogically aligned reward model that translates instructional validity, learning-process support, and affective support into trajectory-level reward signals.
	
	\item We propose a multi-objective RL framework that balances diverse pedagogical objectives through reward discretization and advantage aggregation.
	
	\item We demonstrate that PEARL achieves high answer correctness and strong pedagogical quality, outperforming tutoring-specific open-source systems, as confirmed by blind human evaluations.

\end{itemize}

\section{Methodology}

\begin{figure*}[t]
	\centering
	\includegraphics[width=0.85\linewidth]{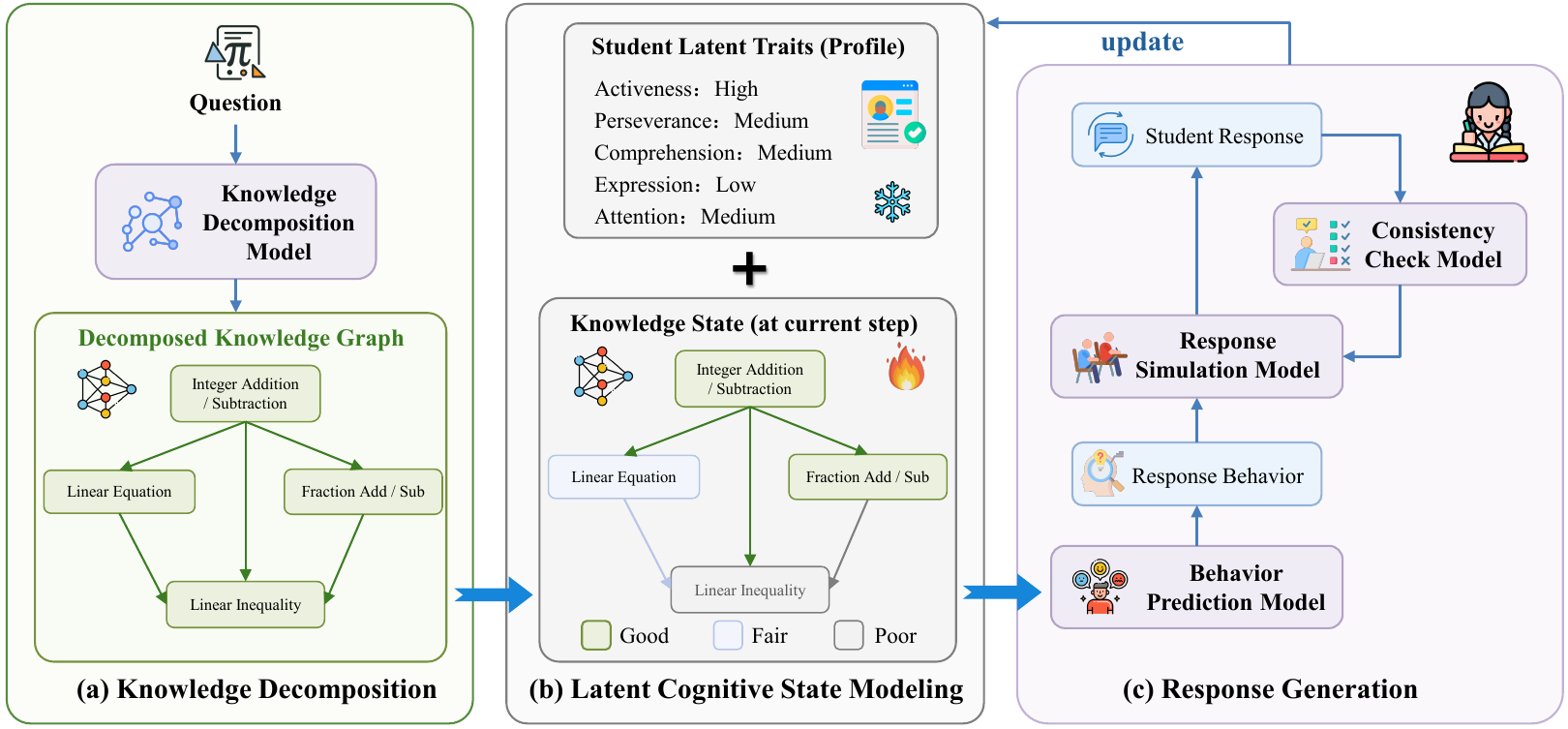}
	\caption{Overview of the controllable student simulator: (a) knowledge decomposition, (b) latent cognitive-state modeling, and (c) response generation through behavior prediction and consistency checking.}
	\label{fig:student_simulator}
\end{figure*}

\subsection{PEARL Overview and Problem Formulation}
We consider the problem of training a Socratic tutoring agent in multi-turn tutoring interactions, as shown in Figure~\ref{fig:multi_agent_interaction}. Unlike standard question answering, Socratic tutoring requires not only producing correct solutions but also delivering progressive guidance that helps students develop understanding step by step. To this end, we formulate PEARL as a closed-loop framework consisting of three components: \textit{controllable student simulation} for interactive environment construction, \textit{pedagogically aligned reward modeling} for fine-grained trajectory evaluation, and \textit{pedagogically aligned tutor optimization} for policy learning.

Formally, given a problem instance $q$, a tutoring episode $\tau$ is represented as a multi-turn trajectory
\begin{align}
	\tau = (q, u^1, a^1, u^2, a^2, ..., u^t, a^t, ..., u^n, a^n),
\end{align}
where $u^t$ and $a^t$ are the student and tutor utterances at turn $t$, respectively, and $n$ is the total number of turns. After the dialogue ends, the full trajectory $\tau$ is evaluated by a judge model $\pi_{\text{judge}}$, which produces a $K$-dimensional reward vector
\begin{align}
	\mathbf{r}(\tau) = [r_1(\tau), r_2(\tau), ..., r_K(\tau)].
\end{align}
PEARL then optimizes the tutor policy $\pi_\theta$ to achieve strong performance across these objectives.

The remainder of this section details the three components of controllable student simulation (Section~\ref{student_agent}), pedagogically aligned reward modeling (Section~\ref{judge_agent}), and pedagogically aligned tutor optimization (Section~\ref{tutor_agent}).




\subsection{Controllable Student Simulation}
\label{student_agent}
Effective tutor training requires interaction with students of varying abilities, misconceptions, and learning behaviors. Existing LLM simulators often encode these properties jointly in a prompt, limiting independent control of knowledge and behavioral tendencies. We instead introduce a structured simulation process that models student knowledge and engagement tendencies, then conditions behavioral planning and response generation on these states. As shown in Figure~\ref{fig:student_simulator}, the process includes knowledge decomposition, latent cognitive-state modeling, and response generation.

\paragraph{Knowledge Decomposition.}
Given a problem $q$ and its reference solution, an LLM extracts and normalizes the required knowledge units. We then construct a directed knowledge graph using fixed prerequisite rules, where $v_i \rightarrow v_j$ indicates that mastering $v_i$ is necessary to apply $v_j$. These rules are derived from standard mathematical curricula and common solution dependencies and are fixed across datasets. This network provides an explicit representation of the knowledge structure required for solving the problem. For a student at turn $t$, we denote the mastery states over these units as $\mathbf{m}^t$. By assigning different states to students, the simulator can explicitly control variations in knowledge mastery.

\paragraph{Latent Cognitive State Modeling.}
Beyond what the student knows, effective simulation also requires modeling how the student engages with tutoring. We therefore introduce a latent cognitive profile $\mathbf{c}$ to characterize student-level cognitive and behavioral tendencies, including activeness, learning perseverance, comprehension ability, expressive ability, and attention. During multi-turn interactions, the mastery state $\mathbf{m}^t$ is dynamically updated according to the problem, dialogue context, tutor action, and student response:
\begin{align}
	\mathbf{m}^{t+1} = f_{\text{update}}(q, \mathbf{h}^{t-1}, \mathbf{m}^t, a^t, u^t).
\end{align}

\paragraph{Response Generation.}
Given the problem $q$, dialogue history $\mathbf{h}^{t-1} = (u^1, a^1, ..., u^{t-1}, a^{t-1})$, the current mastery states $\mathbf{m}^t$, and the latent cognitive profile $\mathbf{c}$, the simulator generates student responses through cognition-conditioned planning and realization. It first predicts a latent behavioral intent $\hat{z}^t$ using the planning model $\pi_{\text{plan}}$, and then generates a natural-language response $u^t$ conditioned on this intent, knowledge mastery, and cognitive profile. The response is further refined to better align with the predicted intent:
\begin{gather}
	\hat{z}^t \sim \pi_{\text{plan}}(\cdot \mid q, \mathbf{h}^{t-1}, \mathbf{m}^t, \mathbf{c}), \\
	u^t \sim \pi_{\text{stu}}(\cdot \mid q, \mathbf{h}^{t-1}, \mathbf{m}^t, \mathbf{c}, \hat{z}^t).
\end{gather}
This design enables the simulator to generate profile-conditioned behaviors, including correct attempts, partial understanding, misconceptions, help-seeking, and off-task responses, while exposing the assumed mastery and profile variables used to generate them.

Overall, explicitly representing simulation state before response generation provides a controllable and behaviorally diverse interaction environment for training Socratic tutoring agents. Simulator realism and transfer are evaluated separately in Section~\ref{sec:experiments_reorganized}.

\begin{figure*}[t]
	\centering
	\includegraphics[width=1.0\linewidth]{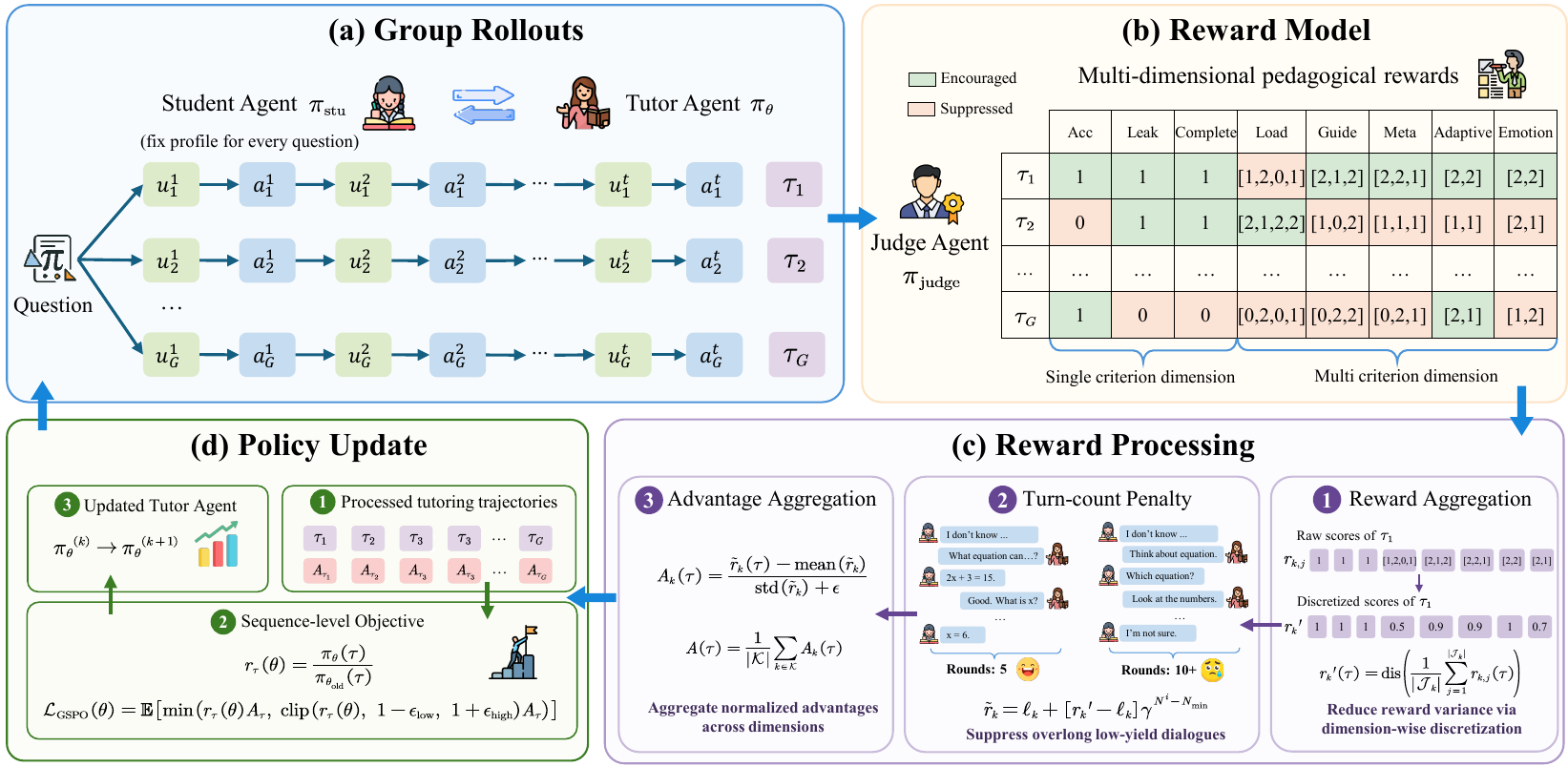}
	\caption{PEARL's tutor optimization framework, comprising (a) group rollouts, (b) pedagogically aligned multidimensional evaluation with a generative reward model, (c) reward processing with aggregation, turn-count penalty, and advantage aggregation, and (d) sequence-level policy update for optimizing the tutor agent.}
	\label{fig:tutor_optimization}
\end{figure*}

\subsection{Pedagogically Aligned Reward Modeling}
\label{judge_agent}
After constructing a controllable interaction environment, training a Socratic tutor requires signals that capture not only final correctness but also multi-turn guidance, answer-leakage avoidance, and pedagogical scaffolding \citep{team2025evaluating, team2024learnlm}. Drawing on learning-science-based rubrics for LLM-based tutoring, we design a pedagogically aligned reward model that translates pedagogical principles into executable, fine-grained trajectory-level criteria. We add task-grounded dimensions for accuracy, leakage control, and completeness, and condition evaluation on cognitive state.

Given a complete tutoring trajectory $\tau$, the judge model $\pi_{\text{judge}}$ evaluates the tutor's performance conditioned on the problem, reference solution, and student cognitive states reflected throughout the interaction. Conditioning on cognitive states allows the judge to assess whether the tutor's behavior is pedagogically appropriate for the student's evolving understanding, rather than evaluating the dialogue independently of the student's current understanding. We define the reward over several pedagogical dimensions, including tutoring accuracy, answer-leakage control, tutoring completeness, cognitive load management, Socratic guidance, metacognitive facilitation, adaptive teaching, and emotional support. Together, these dimensions capture three complementary aspects of tutoring quality: instructional correctness, pedagogical process quality, and affective support. Formally, let
\begin{equation}
	\begin{aligned}
		\mathcal{K} = \{ & \mathrm{acc}, \mathrm{leak}, \mathrm{complete}, \mathrm{load}, \\
		& \mathrm{guide}, \mathrm{meta}, \mathrm{adaptive}, \mathrm{emotion}\},
	\end{aligned}
	\label{eq:rm-multi-dimension}
\end{equation}
denote the set of reward dimensions. For each dimension $k \in \mathcal{K}$, we define a set of dimension-specific scoring criteria $\mathcal{J}_k$, where each criterion $j \in \mathcal{J}_k$ outputs a discrete score $\hat{r}_{k,j}(\tau)$. Since incomplete tutoring trajectories should not receive positive pedagogical rewards, we first derive a binary completion indicator $g(\tau) \in \{0,1\}$ from the completion dimension, where $g(\tau)=0$ indicates that the tutoring process is incomplete. We then gate each criterion-level reward by
\begin{equation}
	r_{k,j}(\tau) = \hat{r}_{k,j}(\tau) \cdot g(\tau), \quad \forall k \in \mathcal{K}.
\end{equation}
Detailed definitions of evaluation dimensions are provided in Section~\ref{sec_appendix:evaluation_metric}.

\subsection{Pedagogically Aligned Tutor Optimization}
\label{tutor_agent}

\subsubsection{Cold-Start Supervised Fine-Tuning}
Before reinforcement learning, we introduce a cold-start supervised fine-tuning stage to initialize the tutor policy. This stage equips the model with fundamental tutoring skills, such as progressive prompting and step-by-step guidance, thereby providing a pedagogical prior for subsequent RL.

Specifically, we first perform SFT on the trainable policy model $\tilde{\pi}_{\theta}$ using the dataset $\mathcal{D}$, which consists of dialogue data generated by DeepSeek-V3.2 as the tutor agent.
\begin{align}
	\mathcal{L}^{\text{SFT}}(\theta)
	= \mathbb{E}_{\tau \sim \mathcal{D}}
	\Big[ -\sum_{t=1}^{n}\log \tilde{\pi}_{\theta}(a^t \mid q, h^t) \Big],
\end{align}
where $n$ denotes the number of tutoring turns in trajectory $\tau$. The resulting policy is used to initialize the RL stage.

\subsubsection{Multi-objective Reinforcement Learning}
Building on the multidimensional rewards from Section~\ref{judge_agent}, we use RL to align the tutor policy with pedagogical principles. Unlike conventional RL \citep{li2025torl, chang2026thor}, Socratic tutoring requires optimizing multiple pedagogical objectives. Directly averaging the corresponding dimension-level rewards into a single terminal reward can destabilize training and cause high-variance objectives to dominate optimization \citep{song2025cultivating, liu2026gdpo}. To address this, we propose a pedagogically aligned multi-objective RL method based on reward and advantage aggregation, as shown in Figure~\ref{fig:tutor_optimization}.

For each problem $q$, we fix the student's initial mastery state and cognitive profile within the group and use the initialized tutor policy $\pi_\theta$ to interact with the student simulator, obtaining a group of tutoring trajectories $\{\tau^i\}_{i=1}^{G}$ under consistent evaluation conditions. The reward model $\pi_{\mathrm{judge}}$ evaluates each trajectory and produces criterion-level rewards $\{r_{k,j}(\tau^i)\}$ across multiple pedagogical dimensions.

\begin{table*}[t]
	\centering
	\small
	\renewcommand{\arraystretch}{1.0}
	\setlength{\tabcolsep}{3.5pt}
		\begin{tabular}{lcccccccccc}
			\toprule[1.5pt]
			Model & Acc & Leak & Complete & Load & Guide & Meta & Adaptive & Emotion & Avg. & Rounds \\
			\midrule
			\rowcolor{gray!8} \multicolumn{11}{c}{\textit{Closed-source LLMs}} \\
			GPT-5.1 & 94.1 & 51.6 & 95.2 & 94.2 & 85.5 & 80.6 & 90.0 & 94.6 & 85.7 & 6.2 \\
			Gemini-3-Flash & 97.1 & 80.4 & 98.6 & 98.2 & 95.4 & 57.8 & 97.3 & 98.3 & 90.4 & 6.0 \\
			
			\midrule
			\rowcolor{gray!8} \multicolumn{11}{c}{\textit{Open-source LLMs}} \\
			DeepSeek-V3.2 & 81.6 & 61.6 & 94.1 & 89.6 &	84.3 & 45.1 & 80.9 & 91.6 & 78.6 & 6.8 \\
			Kimi-K2-Instruct & 76.1 & 44.6 & 87.4 &	85.8 & 73.2 & 33.7 & 71.6 & 83.4 & 69.5 & 5.2 \\
			Qwen3-235B-A22B & 83.7 & 32.2 &	93.2 & 86.5 & 68.3 & 35.8 &	72.0 & 92.4 & 70.5 & 5.2 \\
			MuduoLLM-14B & 58.8 & 33.6 & 72.4 & 58.4 & 50.4 & 31.3 & 47.2 & 69.4 & 52.7 & 5.7 \\
			InnoSpark-plus-72B & 58.1 & 22.1 & 67.6 & 50.5 & 38.7 & 34.0 & 40.4 & 64.9 & 47.1 & 5.2 \\
			
			\midrule
			\rowcolor{gray!8} \multicolumn{11}{c}{\textit{Ours}} \\
			Qwen3-30B-A3B (Baseline) & 85.7 & 41.1 & 92.2 & 72.3 & 73.1 & 37.0 & 60.9 & 90.7 & 69.1 & 5.3 \\
			\rowcolor{green!6}
			PEARL-30B (Ours) & 88.6 \raisebox{0.8ex}{\scriptsize\textcolor{green!70!black}{$\uparrow$ 2.9}} & 76.5\raisebox{0.8ex}{\scriptsize\textcolor{green!70!black}{$\uparrow$ 35.4}} & 94.7\raisebox{0.8ex}{\scriptsize\textcolor{green!70!black}{$\uparrow$ 2.5}} & 86.9\raisebox{0.8ex}{\scriptsize\textcolor{green!70!black}{$\uparrow$ 14.6}} & 77.3\raisebox{0.8ex}{\scriptsize\textcolor{green!70!black}{$\uparrow$ 4.2}} & 82.6\raisebox{0.8ex}{\scriptsize\textcolor{green!70!black}{$\uparrow$ 45.6}} & 73.5\raisebox{0.8ex}{\scriptsize\textcolor{green!70!black}{$\uparrow$ 12.6}} & 93.3\raisebox{0.8ex}{\scriptsize\textcolor{green!70!black}{$\uparrow$ 2.6}} & 84.2\raisebox{0.8ex}{\scriptsize\textcolor{green!70!black}{$\uparrow$ 15.1}} & 6.9 \\
			
			\bottomrule[1.5pt]
		\end{tabular}
	\caption{Comparison with state-of-the-art methods on tutoring evaluation. PEARL achieves strong performance against both open-source and proprietary LLMs.}
	\label{exp:external_evaluator}
\end{table*}

\paragraph{Reward Aggregation.}
Since criteria within a dimension reflect the same pedagogical objective, we aggregate criterion-level scores into a dimension-level reward. However, this aggregation converts discrete criterion-level scores into continuous values, making direct advantage computation sensitive to minor fluctuations. To mitigate this, we discretize the aggregated reward as follows:
\begin{equation}
	r_k'(\tau^i) = \mathrm{dis}( \frac{1}{|\mathcal{J}_k|} \sum_{j=1}^{|\mathcal{J}_k|} r_{k,j}(\tau^i) ),
\end{equation}
where $\mathrm{dis}(\cdot)$ maps each continuous reward in $[0,1]$ to the nearest value in $\{0.1, 0.3, 0.5, 0.7, 0.9\}$. Each discrete value corresponds to a reward bin, and we denote the lower boundary of the bin assigned to $r_k'(\tau^i)$ as $\ell_k^i$. This preserves the coarse-grained quality ordering within each dimension while improving reward stability.

\paragraph{Turn-count Penalty.}
We further introduce a turn-count penalty to discourage unnecessarily long tutoring dialogues. Let $N^i$ denote the total number of turns in trajectory $\tau^i$, and let $N_{\min}$ denote the minimum number of turns among completed trajectories in the same group. The penalized reward is then defined as
\begin{gather}
	\tilde{r}_k(\tau^i) = \ell_k^i + \big[ r_k'(\tau^i) - \ell_k^i \big] \gamma^{N^i - N_{\min}},
\end{gather}
where $\gamma \in (0,1]$ is a decay factor. By penalizing only the portion above the bin lower boundary, this design discourages inefficiently long dialogues while preserving the coarse-grained reward hierarchy induced by discretization.


\paragraph{Advantage Aggregation.}
We compute a normalized advantage for each reward dimension across the sampled group:
\begin{equation}
	A_k(\tau^i) = \frac{\tilde{r}_k(\tau^i) - \mathrm{mean} ( \{ \tilde{r}_k(\tau^{i'}) \}_{i'=1}^{G} ) }{\mathrm{std}( \{ \tilde{r}_k(\tau^{i'}) \}_{i'=1}^{G} ) + \epsilon},
\end{equation}
where $\epsilon$ ensures numerical stability. The final trajectory-level advantage is obtained by averaging across dimensions:
\begin{equation}
	A(\tau^i) = \frac{1}{|\mathcal{K}|} \sum_{k \in \mathcal{K}} A_k(\tau^i).
\end{equation}
This design aggregates and discretizes rewards within each dimension, then normalizes and aggregates advantages across dimensions. It places pedagogical objectives on comparable scales, limiting domination by high-variance dimensions and stabilizing sparse or low-variance rewards. By default, normalized advantages are weighted uniformly. More generally, the final advantage can be written as $A(\tau^i)=\sum_{k\in\mathcal{K}}w_k(\tau^i)A_k(\tau^i)$, where weights sum to one and may depend on simulated student state.

We then optimize the policy $\pi_\theta$ using GSPO \citep{zheng2025group}, which operates at the sequence level rather than the token level, thereby alleviating the instability of GRPO in MoE training. The overall optimization objective is
\begin{equation}
	\begin{split}
		\mathcal{J} & ^{\text{RL}}_{\pi_{\theta}}(\theta) = \mathbb{E}_{[ q \sim \mathcal{D}_{\text{RL}}, \{\tau^i\}_{i=1}^G \sim \pi_{\theta}(\cdot|q) ]} \Big[ \frac{1}{G} \sum_{i=1}^G \min  \\
		& \big( s_i A(\tau^i), \text{clip} ( s_i, 1 - \varepsilon_{\text{low}}, 1+\varepsilon_{\text{high}} ) A(\tau^i) \big) \Big],
	\end{split}
	\label{eq:gspo_objective}
\end{equation}
where $G$ is the group size, $\varepsilon_{\text{low}}$ and $\varepsilon_{\text{high}}$ are the lower and upper clipping ratios, and the sequence-level importance ratio $s_i$ is computed over tutor-generated tokens:
\begin{equation}
	s_i = \big(\frac{\pi_{\theta}(\tau^{i}|q)}{\pi_{{\theta}_{\text{old}}} (\tau^{i}|q)} \big)^{\frac{1}{|\tau^{i}|}}.
\end{equation}
Notably, our multi-objective advantage design is generally applicable to value-model-free policy gradient methods, such as DAPO, GSPO, and SAPO.

\section{Experiments}
\label{sec:experiments_reorganized}

\subsection{Experimental Setup}

\paragraph{Benchmarks.}
We evaluate PEARL on GSM8K \citep{cobbe2021training} for middle-school mathematics, MATH-500 \citep{hendrycks2021measuring} for high-school mathematics, and the education-oriented benchmarks MathTutorBench \citep{macina2025mathtutorbench} and MathDial \citep{macina2023mathdial}. Together, they cover diverse mathematical domains and both problem-solving and tutoring contexts.

\begin{table*}[t]
	\centering
	\small
	\renewcommand{\arraystretch}{1.0}
	\setlength{\tabcolsep}{4pt}
		\begin{tabular}{l|cccccc|cccccccccc}
			\toprule[1.5pt]
			\multirow{2}{*}{}
			& Cold & Ctrl. Stu. & Epis. & Adv. & \multirow{2}{*}{Dis.} & Turn
			& \multirow{2}{*}{Acc} & \multirow{2}{*}{Leak}
			& \multirow{2}{*}{Complete} & \multirow{2}{*}{Load}
			& \multirow{2}{*}{Guide} & \multirow{2}{*}{Meta}
			& \multirow{2}{*}{Adaptive} & \multirow{2}{*}{Emotion}
			& \multirow{2}{*}{Avg.} & \multirow{2}{*}{Rounds} \\
			& Start & Sim. & RL & Agg. & & Pen.
			& & & & & & & & & & \\
			
			\midrule
			T1 \quad & & & & & & & 87.9 & 63.8 & 93.1 & 89.3 & 79.1 & 50.0 & 81.2 & 92.6 & 79.6 & 5.3 \\
			T2 & \checkmark & & & & & & 79.7 & 77.4 & 87.4 & 85.5 & 82.2 & 55.4 & 81.3 & 86.5 & 79.4 & 6.9 \\
			T3 & \checkmark & \checkmark & & & & & 83.5 & 75.9 & 90.0 & 87.2 & 84.2 & 57.9 & 83.9 & 89.4 & 81.5 & 6.8 \\
			T4 & \checkmark & \checkmark & \checkmark & & & & 87.3 & 74.1 & 94.1 & 86.6 & 91.2 & 90.8 & 90.5 & 94.1 & 88.6 & 16.7 \\
			T5 & \checkmark & \checkmark & \checkmark & \checkmark & & & 86.6 & 84.7 & 93.6 & 91.7 & 92.8 & 92.4 & 92.3 & 93.6 & 91.0 & 16.4 \\
			T6 & \checkmark & \checkmark & \checkmark & \checkmark & \checkmark & & 88.4 & 84.7 & 95.0 & 92.6 & 93.3 & 94.0 & 93.9 & 95.0 & 92.1 & 15.0 \\
			\rowcolor{green!6}
			T7 & \checkmark & \checkmark & \checkmark & \checkmark & \checkmark & \checkmark & 94.0 & 89.3 & 96.1 & 94.4 & 88.6 & 95.0 & 90.7 & 95.1 & \textbf{92.9} & 6.9 \\
			\bottomrule[1.5pt]
		\end{tabular}
	\caption{Component ablation of PEARL. T1--T7 progressively add different components. Cold Start denotes cold-start supervised fine-tuning; Ctrl. Stu. Sim. denotes the controllable student simulator; Epis. RL denotes episodic reinforcement learning; Adv. Agg. denotes advantage aggregation; Dis. denotes reward discretization; and Turn Pen. denotes the turn-count penalty.}
	
	\label{exp:main_ablation}
\end{table*}

\paragraph{Compared Tutors and Fair Comparison.}
We compare PEARL with its Qwen3-30B-A3B base model \citep{yang2025qwen3}, proprietary general-purpose models (GPT-5.1 and Gemini-3-Flash), open-source general-purpose models including DeepSeek-V3.2 \citep{liu2025deepseek}, Kimi-K2-Instruct \citep{team2025kimi}, and Qwen3-235B-A22B \citep{yang2025qwen3}, and tutoring-oriented systems MuduoLLM-14B \citep{MuduoLLM2025} and InnoSpark-plus-72B \citep{song2025cultivating}. General-purpose models receive the same model-agnostic Socratic instruction, while tutoring-specific systems use their recommended configurations. All tutors are evaluated with matched problems, initial student profiles, mastery states, simulator conditions, stopping criteria, and generation limits. Complete prompts, checkpoints, inference settings, and the fair-comparison protocol are provided in Appendix~\ref{sec_appendix:implementation_details}.


\begin{table}[t]
	\centering
	\small
	\renewcommand{\arraystretch}{1.0}
		\begin{tabular}{lcccc}
			\toprule[1.3pt]
			Compared Tutor & Win & Tie & Loss & Preference \\
			\midrule
			GPT-5.1 & 38.32 & 9.24 & 52.45 & 42.94 \\
			Gemini-3-Flash & 38.10 & 6.10 & 55.80 & 41.15 \\
			InnoSpark-plus-72B & 81.49 & 3.66 & 14.86 & 83.31 \\
			MuduoLLM-14B & 77.74 & 3.57 & 18.69 & 79.52 \\
			Qwen3-30B-A3B & 64.10 & 4.73 & 31.17 & 66.47 \\
			\bottomrule[1.3pt]
		\end{tabular}
	\caption{Blind pairwise evaluation of PEARL against other tutors by human teachers. Preference is computed as $\text{Win}+0.5\times\text{Tie}$, with values above 50\% favoring PEARL.}
	\label{exp:human_tutor_preference}
\end{table}

\subsection{Comparison with State-of-the-Art Methods}

\paragraph{Evaluation Protocol.}
We evaluate end-to-end tutoring quality through external evaluation and blind teacher comparisons. Gemini-3.1-Pro scores complete tutoring trajectories on accuracy (Acc), answer-leakage control (Leak), instructional completeness (Complete), cognitive-load management (Load), Socratic guidance (Guide), metacognition (Meta), adaptivity (Adaptive), and emotional support (Emotion). Experienced teachers also conduct blind pairwise comparisons of complete dialogues, with model identities concealed and presentation order randomized.

\paragraph{External-Evaluator Results.}
As shown in Table~\ref{exp:external_evaluator}, PEARL achieves an average external-evaluator score of 84.2, compared with 69.1 for its Qwen3-30B-A3B base model. The largest gains appear in Leak, which increases from 41.1 to 76.5, and Meta, which rises from 37.0 to 82.6, while the average number of rounds increases moderately from 5.3 to 6.9. PEARL also outperforms MuduoLLM-14B and InnoSpark-plus-72B, although it remains behind GPT-5.1 and Gemini-3-Flash in the overall external score.

\paragraph{Blind Teacher Comparison.}
In addition to the LLM-based evaluation, we conduct an expert human evaluation. Table~\ref{exp:human_tutor_preference} presents the results of pairwise comparisons between PEARL and other models, reporting the Win, Tie, Loss, and Preference rates. We compute the preference rate as $\text{Preference}=\text{Win}+0.5\times\text{Tie}$, where a higher value indicates a stronger preference for PEARL. The results show that human teachers prefer PEARL over its base model and the two tutoring-oriented systems, with preference rates ranging from 66.47\% to 83.31\%. By contrast, GPT-5.1 and Gemini-3-Flash are preferred over PEARL. Overall, these findings demonstrate that PEARL substantially improves upon its base model and achieves strong performance among the evaluated open-source tutoring systems.

\subsection{Ablation Study}
To systematically assess the effectiveness of each PEARL component, we construct seven Qwen3-30B-A3B variants, T1--T7, by progressively incorporating the components of PEARL. Table~\ref{exp:main_ablation} reports their configurations, dimension-wise scores, overall averages, and dialogue lengths. The tutoring scores in the ablation study are provided by PEARL-RM.

\textbf{Cold start and student simulation.}
Cold-start SFT with data collected using our controllable simulator improves the average score from 79.6 in T1 to 81.5 in T3. T2 and T3 use the same SFT procedure but differ in the simulator used to collect tutor--student trajectories: T2 uses the prompt-based SocraticLM simulator \citep{liu2024socraticlm}, whereas T3 uses our controllable simulator. This replacement raises the average from 79.4 to 81.5.

\textbf{Episodic RL and reward processing.}
Episodic RL raises the average from 81.5 in T3 to 88.6 in T4, with notable gains in guidance, metacognition, and adaptivity, but increases dialogue length from 6.8 to 16.7 rounds. Advantage aggregation and reward discretization further improve the score to 91.0 in T5 and 92.1 in T6, supporting more balanced and stable multi-objective optimization. Finally, the turn-count penalty reduces dialogue length from 15.0 rounds in T6 to 6.9 in T7 while increasing the average to 92.9. Despite moderate decreases in guidance and adaptivity, T7 achieves the best overall balance between tutoring quality and efficiency. Training dynamics are reported in Appendix~\ref{sec_appendix:training_curves}.

\begin{table*}[t]
	\centering
	\small
	\renewcommand{\arraystretch}{1.0}
	\setlength{\tabcolsep}{2.5pt}
		\begin{tabular}{lcccccc}
			\toprule[1.5pt]
			Student Simulator & Source ID Acc. $\downarrow$ & Label Coverage $\uparrow$ & Attempt Entropy $\uparrow$ & Unique Label Sets $\uparrow$ & Label-Set Entropy $\uparrow$ & Realism Avg. $\uparrow$ \\
			\midrule
			MathDial Student & 63.3 & 11/13 & 0.562 & 62 & 3.763 & 3.88 \\
			SocraticLM Student & 65.3 & 11/13 & 0.545 & 47 & 3.215 & 3.49 \\
			\rowcolor{green!6}
			Our Simulator & \textbf{61.8} & \textbf{13/13} & \textbf{0.651} & \textbf{81} & \textbf{4.317} & \textbf{4.46} \\
			\bottomrule[1.5pt]
		\end{tabular}
	\caption{Evaluation of student simulators on response distinguishability, behavioral diversity, and trajectory realism. Lower Source ID Acc. is better, while higher values are better for all other metrics.}
	\label{exp:main_simulator_validity}
\end{table*}

\subsection{Reward-Model Validity}

Reward models for tutoring differ in their rubrics and training objectives \citep{macina2025mathtutorbench, song2025cultivating}, making agreement with automatic annotations alone insufficient. We therefore evaluate them on 300 randomly sampled dialogue pairs, independently compared by two experienced teachers before disagreement resolution, and report agreement by comparison margin.

\begin{table}[t]
	\centering
	\small
	\renewcommand{\arraystretch}{1.0}
	\setlength{\tabcolsep}{3pt}
		\begin{tabular}{lcccc}
			\toprule[1.5pt]
			\multirow{2}{*}{Reward Model} & Low & Medium & High & \multirow{2}{*}{Overall} \\
			& Margin & Margin & Margin &  \\
			
			\midrule
			GPT-5.1 & 68.0 & 71.6 & 90.2 & 78.3 \\
			Gemini-3-Flash & 64.0 & 71.6 & 89.4 & 77.0 \\
			Qwen3-32B & 53.3 & 50.0 & 59.3 & 54.7 \\
			Qwen3-235B-A22B & 56.0 & 65.7 & 89.4 & 73.0 \\
			MathTutorBench RM & 45.3 & 54.9 & 50.4 & 50.7 \\
			InnoSpark-HPC-RM-32B & 40.0 & 45.1 & 50.4 & 46.0 \\
			\rowcolor{green!6}
			PEARL-RM (Ours) & \textbf{69.3} & \textbf{76.5} & \textbf{94.3} & \textbf{82.0} \\
			\bottomrule[1.5pt]
		\end{tabular}
	\caption{Agreement of different reward models with teacher preferences on 300 randomly sampled dialogue pairs, stratified by comparison margin.}
	\label{exp:main_rm_agreement}
\end{table}

As shown in Table~\ref{exp:main_rm_agreement}, PEARL-RM achieves the highest agreement overall (82.0\%) and across all margin ranges. Agreement rises from 69.3\% on low-margin pairs to 94.3\% on high-margin pairs, indicating that pedagogically ambiguous comparisons are more challenging. Additional dimension-wise results are provided in Appendix~\ref{sec_appendix:full_ablation}.

\subsection{Student-Simulator Validity}

We evaluate the simulator from three complementary perspectives: ability-conditioned controllability, behavioral diversity and trajectory plausibility, and transfer of the trained tutor to alternative student simulators. The alternatives are the student components used by MathDial \citep{macina2023mathdial} and SocraticLM \citep{liu2024socraticlm}.

\begin{figure}[t]
	\centering
	\includegraphics[width=0.6\linewidth]{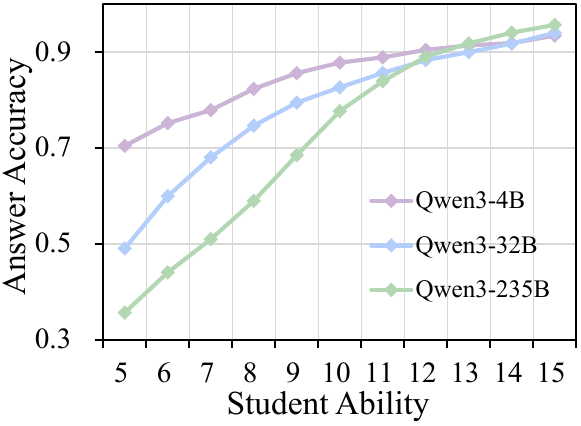}
	\caption{Response accuracy across student ability levels, demonstrating controllable ability-dependent behavior.}
	\label{fig:main_student_controllability}
\end{figure}

\begin{table}[t]
	\centering
	\small
	\renewcommand{\arraystretch}{1.0}
		\begin{tabular}{lccc}
			\toprule[1.5pt]
			Student Simulator & Baseline & PEARL & Gain \\
			\midrule
			MathDial Student & 82.3 & 93.3 & $+11.0$ \\
			SocraticLM Student & 78.6 & 93.5 & $+14.9$ \\
			Our Simulator & 79.6 & 92.9 & $+13.3$ \\
			\bottomrule[1.5pt]
		\end{tabular}
	\caption{Cross-simulator generalization. Scores are average tutoring-evaluation scores under each student simulator.}
	\label{exp:main_cross_simulator}
\end{table}

\paragraph{Ability-Conditioned Controllability.}
Under identical problem settings, simulated students with different cognitive profiles interact with the same tutor. We group the five profile dimensions into low-, medium-, and high-ability levels, with aggregate scores from 5 to 15, and use Gemini-3.1-Pro to evaluate response accuracy. As shown in Figure~\ref{fig:main_student_controllability}, accuracy increases with the configured ability level, demonstrating ability-conditioned controllability.

\paragraph{Turing-Style Distinguishability.}
Table~\ref{exp:main_simulator_validity} evaluates response distinguishability, behavioral diversity, and trajectory plausibility. For Source ID Acc., we sample 1,000 authentic Eedi tutoring contexts \citep{zent2025piivot} truncated before the student response and have each simulator generate a response from the same context. Blinded human experts identify whether each response is authentic or simulated. Because chance accuracy is 50\%, lower accuracy indicates greater similarity to real student responses.

\paragraph{Behavioral Diversity.}
For behavioral diversity, we collect 500 matched dialogues per simulator and use Gemini-3.1-Pro to annotate each response with one attempt-outcome label and zero or more labels from a 13-category behavior taxonomy. Label Coverage counts how many behavior categories appear; Attempt Entropy measures the normalized diversity of attempt outcomes; and Unique Label Sets and Label-Set Entropy measure the variety and dispersion of complete multi-label behavior combinations. Additional frequency-threshold statistics are reported in Appendix~\ref{sec_appendix:simulator_analysis}.

\paragraph{Trajectory Realism.}
For Realism Avg., we design evaluator criteria covering linguistic, contextual, behavioral, and cognitive aspects of student trajectories, and use Gemini-3.1-Pro to score each trajectory on a 1--5 scale. Higher scores indicate more realistic simulated trajectories, with detailed scoring criteria and full breakdowns provided in Appendix~\ref{sec_appendix:simulator_analysis}.

\paragraph{Cross-Simulator Validity.}
Table~\ref{exp:main_cross_simulator} further shows that PEARL improves over its base model by 11.0--14.9 points under all three simulators. This supports cross-simulator robustness in simulated environments.

\section{Conclusion}
We present PEARL, a framework for training Socratic tutors through controllable student simulation, trajectory-level pedagogical reward modeling, and multi-objective RL. The simulator represents diverse student mastery states and cognitive profiles; the reward model jointly evaluates correctness, pedagogical process, and affective support; and the optimization method balances multiple objectives while improving dialogue efficiency. Across four mathematics benchmarks, PEARL substantially improves upon its base model. These gains are consistently observed under an independent external evaluator, in blind teacher comparisons with the base model and tutoring-specific open-source systems, and in evaluations using alternative student simulators. Overall, the results demonstrate improved tutoring-dialogue quality and cross-simulator robustness in simulated environments.

\bibliography{aaai2027}


\appendix

\section{Limitations and Future Work}
This work has several limitations. First, although the proposed student simulator improves controllability and interaction fidelity, it cannot fully capture the complexity of real students in authentic educational settings. In particular, the current framework focuses on short-horizon tutoring interactions and does not explicitly model long-term learning dynamics. In future work, we plan to explore the long-term evolution of student behavior and develop memory-based mechanisms for long-horizon tutoring. Second, our experiments mainly focus on mathematical tutoring, and the generalizability of PEARL to other educational domains remains to be validated. We plan to extend the framework to a broader range of subjects and tasks, such as language learning, physics, and chemistry. Third, although our generative reward model shows strong agreement with expert preferences, it is trained from model-generated annotations and may inherit biases from the annotator model. Future work should incorporate broader human annotations and examine whether reward improvements translate into long-term learning gains for real students.

\begin{table*}[t]
	\centering
	\small
		\begin{tabular}{lccccccccc}
			\toprule[1.5pt]
			Model & Acc & Leak & Complete & Load & Guide & Meta & Adaptive & Emotion & Avg. \\
			\midrule
			GPT-5.1 & 93.4 & 84.9 & 94.8 & 94.6 & 91.1 & 89.0 & 92.0 & 94.6 & 91.8 \\
			Gemini-3-Flash & 95.4 & 93.9 & 97.9 & 97.7 & 96.4 & 74.8 & 97.5 & 97.8 & 93.9 \\
			MuduoLLM-14B & 59.0 & 46.2 & 73.5 & 66.6 & 54.5 & 42.6 & 57.2 & 73.8 & 59.2 \\
			InnoSpark-turbo-7B & 41.7 & 29.9 & 65.6 & 51.7 & 40.9 & 39.4 & 42.2 & 66.4 & 47.2 \\
			InnoSpark-plus-72B & 56.4 & 37.0 & 67.9 & 57.6 & 40.7 & 40.9 & 45.7 & 68.8 & 51.9 \\
			DeepSeek-V3.2 & 89.8 & 76.6 & 93.7 & 91.6 & 87.5 & 57.4 & 87.5 & 93.1 & 84.7 \\
			Kimi-K2-Instruct & 83.4 & 66.7 & 87.7 & 87.0 & 77.6 & 43.1 & 79.2 & 86.5 & 76.4 \\
			GLM-4.7 & 82.7 & 84.3 & 86.2 & 85.6 & 81.3 & 52.4 & 82.6 & 86.5 & 80.2 \\
			MiniMax-M2.1 & 89.5 & 59.5 & 94.0 & 92.4 & 81.1 & 58.0 & 83.8 & 93.8 & 81.5 \\
			Qwen3-235B-A22B & 89.6 & 56.3 & 95.4 & 90.2 & 75.4 & 49.9 & 81.4 & 95.0 & 79.2 \\
			Qwen3-30B-A3B (Baseline) & 87.9 & 63.8 & 93.1 & 89.3 & 79.1 & 50.0 & 81.2 & 92.6 & 79.6 \\
			\textbf{PEARL-30B (Ours)} & 94.0 & 89.3 & 96.1 & 94.4 & 88.6 & 95.0 & 90.7 & 95.1 & 92.9 \\
			\bottomrule[1.5pt]
		\end{tabular}
	\caption{Full dimension-wise evaluation by PEARL-RM.}
	\label{tab:appendix_pearl_rm_full}
\end{table*}

\section{Related Work}
LLMs have shown strong potential in education. In this section, we review three lines of work most relevant to training effective tutoring agents: student simulation, tutoring dialogue evaluation, and pedagogical alignment for LLM tutors.

\subsection{LLM Tutors and Student Simulation}
Recent studies have shown that LLMs hold strong promise for personalized learning. Early work mainly relied on prompt engineering to elicit Socratic tutoring behaviors \citep{park2024empowering, liu2024socraticlm, zhang2024spl}, demonstrating their ability to provide both answers and multi-turn tutoring. More recent systems further emphasize pedagogically grounded interaction, including scaffolded prompting, student-centered dialogue, and multi-turn learning support \citep{dinucu2025problem, scarlatos2025training, shi2025educationq}.

Student simulation is also important for tutor design, as exposure to students with diverse cognitive states helps tutors adapt feedback and interaction strategies to individual learners \citep{villegas2025adaptive}. Prior work has explored simulating learners with different personality traits \citep{liu2024personality}, using LLMs for personalized error correction \citep{daheim2024stepwise}, fine-tuning LLMs on diverse student data \citep{sonkar2024llm, xu2025classroom}, and incorporating learner profiles into tutoring systems \citep{park2024empowering, gao2025agent4edu}. However, most existing methods still rely on end-to-end response generation, which limits controllability. In contrast, we propose a controllable student simulator with cognition-decision decoupling for more faithful and controllable learning dynamics.

\subsection{Tutoring Dialogue Evaluation}
Task-specific evaluation typically relies on fixed question-answer settings to assess domain knowledge and reasoning ability, such as GPQA \citep{rein2024gpqa}, MMLU \citep{hendrycks2020measuring}, MMLU-Pro \citep{wang2024mmlu}, and MATH \citep{hendrycks2021measuring}. Evaluating tutoring ability, however, is more challenging, as it must jointly account for pedagogical quality and task correctness over multi-turn interactions. Early studies, therefore, often relied on human annotation \citep{macina2023mathdial}, which is difficult to scale. With the rise of LLM-as-a-judge \citep{zheng2023judging, gu2024survey}, recent work has introduced automated benchmarks and model-based evaluators for tutoring assessment \citep{macina2025mathtutorbench, maurya2025unifying, wei2025elmes}. \citet{he2024socreval} use Socratic prompting for reference-free reasoning evaluation with LLMs. However, existing methods often remain insufficient for tutor optimization: they provide limited fine-grained, trajectory-level judgment rules, rarely unify pedagogical quality with objective correctness, and are not directly designed to produce reward signals for RL.

\subsection{Pedagogical Alignment for LLM Tutors}
Pedagogical alignment is central to training effective LLM tutors. Existing approaches mainly include supervised fine-tuning \citep{bonino2024euler}, preference optimization (e.g., DPO), and RL methods such as PPO \citep{schulman2017proximal}, GRPO \citep{shao2024deepseekmath}, DAPO \citep{yu2025dapo}, and GSPO \citep{zheng2025group}. In educational settings, \citet{scarlatos2025training} applied DPO to improve student outcomes, \citet{dinucu2025problem} used online RL to balance answer withholding and pedagogical correctness, and \citet{song2025cultivating} used GRPO to optimize dimensions such as helpfulness, personalization, and creativity. \citet{jiang2025evolutionary} combined evolutionary search with RL to optimize Socratic tutoring policies. However, tutoring is inherently multi-objective, and different objectives often exhibit markedly different variance scales and sparsity patterns. Existing methods typically combine them through simple linear averaging \citep{song2025cultivating}, which can cause high-variance objectives to dominate updates and destabilize training, especially in multi-turn dialogues.

\section{Additional Experiments}
\label{sec_appendix:full_results}

\subsection{Training-Aligned PEARL-RM Evaluation}

Given the high cost of external evaluators, we additionally use PEARL-RM as an evaluator and compare our method with other approaches to further validate its effectiveness. The results are reported in Table~\ref{tab:appendix_pearl_rm_full}.

\subsection{Training Curves}
\label{sec_appendix:training_curves}
\begin{figure}[t]
	\centering
	\includegraphics[width=1.0\linewidth]{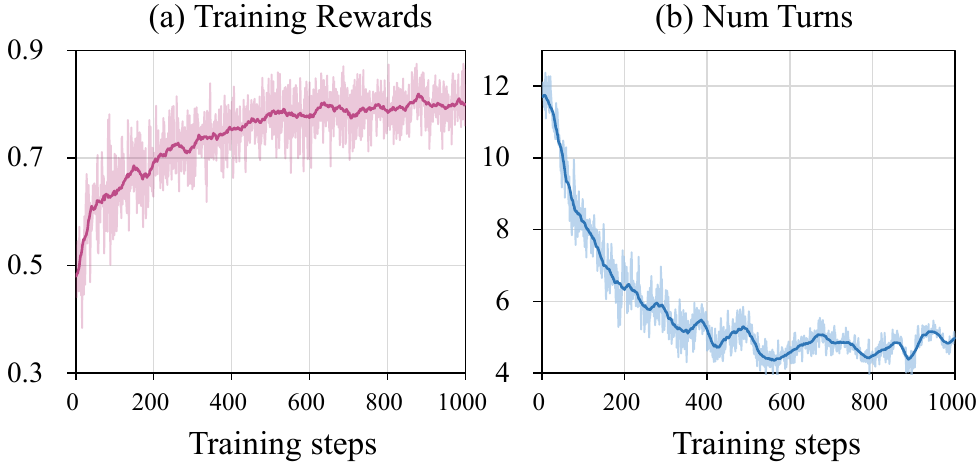}
	\caption{Changes in dialogue turns and within-group rewards during RL. Reward signals improve while the turn-count penalty suppresses unnecessarily long interactions.}
	\label{fig:appendix_rewards_turns}
\end{figure}

We plot the changes in within-group reward and dialogue length over the course of RL training, as shown in Figure~\ref{fig:appendix_rewards_turns}. As training progresses, the within-group reward increases steadily, while the number of dialogue turns gradually decreases.

\subsection{Dimension-Wise Reward-Model Accuracy}
\label{sec_appendix:full_ablation}
To validate the effectiveness of our reward model, we hold out a test split from the reward-model training data, with labels annotated by Gemini-3-Pro. We report the agreement rates for each evaluation dimension on this test set, as shown in Table~\ref{tab:appendix_rm_dimension}.

\begin{table}[t]
	\centering
	\small
	\begin{tabular}{cc}
		\toprule[1.5pt]
		Dimension & Agreement with labels (\%) \\
		\midrule
		Acc & 94.12 \\
		Leak & 89.83 \\
		Complete & 97.49 \\
		Load & 85.79 \\
		Guide & 82.28 \\
		Meta & 79.45 \\
		Adaptive & 80.68 \\
		Emotion & 97.20 \\
		\textbf{Avg.} & \textbf{88.36} \\
		\bottomrule[1.5pt]
	\end{tabular}
	\caption{Agreement with held-out labels from the proprietary annotation source.}
	\label{tab:appendix_rm_dimension}
\end{table}

\section{Student-Simulator Analyses}
\label{sec_appendix:simulator_analysis}

\subsection{Turing-Style Response Distinguishability}
We sample authentic tutoring contexts from the Eedi Question-Anchored Tutoring Dialogues dataset and truncate each dialogue immediately before a real student turn. Given the same preceding context, each student simulator generates a candidate response, which is paired with the corresponding real student response. The two responses are anonymized, and their A/B presentation order is randomized. Evaluators are asked to identify which response was produced by the real student and to independently assess the responses in terms of consistency with the tutor’s preceding message, continuity with the student’s demonstrated cognitive state, and alignment with the student’s language and interaction style. Source-identification accuracy is evaluated against a 50\% chance baseline, with higher values indicating greater residual distinguishability between real and simulated responses. The results are reported in Table~\ref{tab:appendix_turing}.

\begin{table*}[t]
	\centering
	\small
	\begin{tabular}{lcccc}
		\toprule[1.5pt]
		Student Simulator & Correct Attempt & Partially Correct & Incorrect Attempt & No Attempt \\
		\midrule
		MathDial & 72.5 & 21.2 & 3.2 & 3.1 \\
		SocraticLM & 74.1 & 19.6 & 2.2 & 4.1 \\
		Our Simulator & 65.7 & 25.3 & 4.5 & 4.5 \\
		\bottomrule[1.5pt]
	\end{tabular}
	\caption{Attempt-outcome distribution over valid student turns (\%).}
	\label{tab:appendix_attempt_distribution}
\end{table*}

\begin{table*}[t]
	\centering
	\small
	\begin{tabular}{lccc}
		\toprule[1.5pt]
		Behavior & MathDial & SocraticLM & Our Simulator \\
		\midrule
		Misconception & 1.9 & 1.4 & 1.7 \\
		Uncertainty / confusion & 51.8 & 10.4 & 47.6 \\
		Clarification / help-seeking & 8.4 & 2.7 & 11.1 \\
		Self-correction & 5.5 & 3.0 & 3.3 \\
		Persistence / retry & 5.7 & 2.1 & 6.7 \\
		Feedback uptake & 59.3 & 62.7 & 47.0 \\
		Metacognitive monitoring & 87.8 & 57.1 & 54.2 \\
		Positive engagement & 5.1 & 3.2 & 0.5 \\
		Frustration / disengagement & 0.0 & 0.0 & 2.9 \\
		Off-topic / inattention & 0.4 & 0.9 & 1.6 \\
		\bottomrule[1.5pt]
	\end{tabular}
	\caption{Behavior-label frequency over valid turns (\%). Labels are non-exclusive.}
	\label{tab:appendix_behavior_distribution}
\end{table*}

\begin{table*}[t]
	\centering
	\small
	\setlength{\tabcolsep}{5pt}
	\begin{tabular}{lccccccc}
		\toprule[1.5pt]
		Simulator & Coverage & Labels $\geq 1\%$ & Mean Labels & Attempt Entropy & Unique Sets & Label-Set Entropy & Transition Entropy \\
		\midrule
		MathDial & 11/13 & 9 & 2.264 & 0.562 & 62 & 3.763 & 0.946 \\
		SocraticLM & 11/13 & 8 & 1.443 & 0.545 & 47 & 3.215 & 1.031 \\
		Our Simulator & 13/13 & 11 & 1.779 & 0.651 & 81 & 4.317 & 1.275 \\
		\bottomrule[1.5pt]
	\end{tabular}
	\caption{Behavioral-diversity statistics.}
	\label{tab:appendix_behavior_diversity}
\end{table*}

\begin{table}[t]
	\centering
	\small
		\setlength{\tabcolsep}{3.5pt}
		\begin{tabular}{lcc}
			\toprule[1.5pt]
			Comparison & Source ID Acc. $\downarrow$ & $|\mathrm{Acc.}-50|$ $\downarrow$ \\
			\midrule
			Real vs. Our Simulator & 61.8 & 11.8 \\
			Real vs. MathDial Student & 63.3 & 13.3 \\
			Real vs. SocraticLM Student & 65.3 & 15.3 \\
			\bottomrule[1.5pt]
		\end{tabular}
	\caption{Response-source identification (\%).}
	\label{tab:appendix_turing}
\end{table}

\subsection{Behavioral Distribution and Diversity}
We evaluate each simulator on the same set of 500 problems, comprising 250 examples from GSM8K and 250 from MATH-500. We exclude the initial turn, which only presents the problem. For each subsequent valid student response, Gemini-3.1-Pro assigns one attempt-outcome label---correct, partially correct, incorrect, or no attempt---along with zero or more behavioral labels. Because the behavioral annotations are multi-label, their percentages do not necessarily sum to 100\%. Tables~\ref{tab:appendix_attempt_distribution},~\ref{tab:appendix_behavior_distribution}, and~\ref{tab:appendix_behavior_diversity} report the resulting attempt-outcome distributions, the frequencies of selected key behavioral labels, and diversity statistics, respectively.

Our simulator covers all 13 behavior categories and produces the largest attempt, label-set, and transition entropies. MathDial has more labels per turn but lower label-set entropy, indicating frequent co-occurrence of a narrower set of behaviors.

\subsection{Trajectory-Level Plausibility}
Gemini-3.1-Pro rates complete student trajectories on six criteria using a 1--5 scale, with the results shown in Table~\ref{tab:appendix_trajectory_realism}. Linguistic naturalness assesses student-like rather than teacher-like or templated language. Contextual coherence measures direct response to the preceding tutor turn. Behavioral-response plausibility and behavioral continuity assess whether attempts, confusion, help-seeking, correction, retry, and disengagement are locally and temporally reasonable. Cognitive-state consistency evaluates agreement with earlier demonstrated mastery and uncertainty, while plausibility of cognitive change checks that the student absorbs only information made available in the dialogue.

\begin{table*}[t]
	\centering
	\small
	\begin{tabular}{lccccccc}
		\toprule[1.5pt]
		Simulator & Language & Context & Behavior & Behavior Continuity & State Consistency & Cognitive Change & Avg. \\
		\midrule
		MathDial & 3.39 & 4.85 & 4.12 & 3.92 & 3.12 & 3.87 & 3.88 \\
		SocraticLM & 2.80 & 4.62 & 3.66 & 3.19 & 2.97 & 3.68 & 3.49 \\
		Our Simulator & 4.51 & 4.62 & 4.68 & 4.16 & 4.24 & 4.54 & 4.46 \\
		\bottomrule[1.5pt]
	\end{tabular}
	\caption{Trajectory-level plausibility scores evaluated by Gemini-3.1-Pro. Among the simulators compared, our student simulator achieved the highest score.}
	\label{tab:appendix_trajectory_realism}
\end{table*}

\begin{table*}[t]
	\centering
	\small
	\begin{tabular}{lcccccccccc}
		\toprule[1.5pt]
		Tutor & Student Simulator & Acc & Leak & Complete & Load & Guide & Meta & Adaptive & Emotion & Avg. \\
		\midrule
		Qwen3-30B-A3B & SocraticLM & 89.8 & 72.5 & 93.2 & 89.9 & 76.3 & 37.2 & 77.9 & 91.9 & 78.6 \\
		PEARL & SocraticLM & 93.0 & 93.8 & 95.6 & 93.6 & 93.1 & 92.0 & 91.7 & 95.2 & 93.5 \\
		Qwen3-30B-A3B & MathDial & 91.4 & 72.3 & 94.7 & 91.5 & 82.4 & 49.3 & 82.4 & 94.0 & 82.3 \\
		PEARL & MathDial & 92.4 & 93.3 & 95.0 & 93.4 & 93.6 & 93.0 & 91.1 & 94.9 & 93.3 \\
		\bottomrule[1.5pt]
	\end{tabular}
	\caption{Full cross-simulator results. Under SocraticLM and MathDial, PEARL improves over its base model by 14.9 and 11.0 average points, respectively.}
	\label{tab:appendix_cross_simulator}
\end{table*}

\subsection{Cross-Simulator Tutor Evaluation}
To test whether the optimized tutor remains effective under alternative student simulators, we evaluate Qwen3-30B-A3B and PEARL with SocraticLM and MathDial students. The full cross-simulator results are reported in Table~\ref{tab:appendix_cross_simulator}.

\section{Tutor Evaluation Metrics}
\label{sec_appendix:evaluation_metric}

We evaluate tutoring dialogues from the perspectives of both objective correctness and pedagogical quality. Given the problem, reference answer, solution, student profile, and the full tutor-student dialogue, the judge agent assesses the tutor along eight dimensions: \textit{Acc}, \textit{Leak}, \textit{Complete}, \textit{Load}, \textit{Guide}, \textit{Meta}, \textit{Adaptive}, and \textit{Emotion}.

\paragraph{Accuracy (Acc)}
This metric evaluates whether the tutor's teaching process is mathematically and logically correct. The judge checks the correctness of the tutor's explanations, calculations, step-by-step guidance, and feedback to the student. A low score is assigned if the tutor makes factual, computational, or reasoning errors, or fails to guide the student to the reference answer.

\paragraph{Answer-Leakage Control (Leak)}
This metric evaluates whether the tutor \emph{prematurely} reveals the final answer, supplies key calculations, or replaces the student's core reasoning process. Full leakage occurs when the tutor gives the answer or decisive derivation before the student has had an appropriate opportunity to reason. Partial leakage occurs when overly specific hints or candidate answers reduce the task to verification. Open-ended questions, general strategies, conceptual scaffolding, and post-solution summaries are not leakage. Moreover, direct explanation may be pedagogically appropriate after the student has made sufficient attempts, remains persistently stuck, or explicitly requests an explanation. The judge therefore evaluates leakage over the complete trajectory and alongside Complete, Adaptive, and Load, rather than matching answer strings or penalizing every explicit statement.

\paragraph{Completion (Complete)}
This metric evaluates whether the tutor completes the instructional process. The judge checks whether the tutor covers the necessary reasoning steps, addresses student mistakes, helps the student reach the correct answer, and provides an appropriate conclusion. Dialogues that stop too early or miss key steps receive lower scores.

\paragraph{Cognitive Load Management (Load)}
This metric measures whether the tutor presents information in a clear and manageable way. It includes four aspects: whether the explanation is concise, whether the information is clear and easy to understand, whether the response length is appropriate, and whether the tutor avoids irrelevant information.

\paragraph{Socratic Guidance (Guide)}
This metric evaluates whether the tutor promotes active student participation. It includes the quality of the tutor's questioning strategy, the effectiveness of the provided guidance, and whether the tutor encourages the student to remain the main problem solver rather than passively following instructions.

\paragraph{Metacognition (Meta)}
This metric evaluates whether the tutor encourages the student to reflect on their reasoning. It includes whether the tutor guides the student to discover mistakes, prompts the student to explain or verify their thinking, and summarizes the core method or insight after the problem is solved.

\paragraph{Adaptivity (Adaptive)}
This metric evaluates whether the tutor adapts to the student's learning state. It includes whether the tutor can help the student when they are stuck by changing strategies, and whether the tutor adjusts question difficulty according to the student's responses, ability, and demonstrated understanding.

\paragraph{Affective Support (Emotion)}
This metric evaluates whether the tutor creates a positive and supportive learning environment. It considers whether the tutor maintains an encouraging tone and whether student mistakes are handled as learning opportunities rather than failures.

\section{Implementation Details}
\label{sec_appendix:implementation_details}

\subsection{Test Setting}
During inference, we allow at most 20 interaction turns between the tutor agent and the student agent, with a maximum generation length of 8,192 tokens. For both agents, we set the temperature to 1.0, top-$p$ to 1.0, and top-$k$ to 50.

\subsection{Judge Agent}
To improve both interpretability and evaluation reliability, we train the judge model under a generative evaluation paradigm. Specifically, we collect diverse tutoring trajectories generated by different tutor and student models, and use Gemini-3-Pro to annotate each trajectory with dimension-specific scores and corresponding natural-language rationales. After filtering out noisy and inconsistent annotations, we obtain 803K training instances for supervising $\pi_{\text{judge}}$.

We fully fine-tune the judge agent from Qwen3-32B \citep{yang2025qwen3} for 1 epoch, using a global batch size of 256. We adopt AdamW \citep{loshchilov2017decoupled} as the optimizer and fix the learning rate at $5 \times 10^{-6}$. To improve memory efficiency, we employ ZeRO \citep{rajbhandari2020zero} during training. During inference, we set the sampling temperature to 0.3, top-$p$ to 1.0, and top-$k$ to 20.

\subsection{Tutor Agent}
In the cold-start stage, we use DeepSeek-V3.2 \citep{liu2025deepseek} to interact with the student simulator over multiple turns and generate high-quality tutoring dialogues. After applying data filtering strategies such as turn balancing, we construct the cold-start dataset $\mathcal{D}$, which contains 193K samples. Based on this dataset, we conduct full-parameter fine-tuning on Qwen3-30B-A3B-Instruct-2507 \citep{yang2025qwen3} for one epoch, using AdamW with a fixed learning rate of $5 \times 10^{-6}$.

In the reinforcement learning stage, we optimize the tutor model using GSPO. Specifically, the group size $G$ is set to 8, the sampling temperature is set to 1.0, and the learning rate is set to $1 \times 10^{-6}$, without KL regularization. The clipping coefficients are set to $\varepsilon_{\text{high}} = 4 \times 10^{-4}$ and $\varepsilon_{\text{low}} = 3 \times 10^{-4}$. The turn-count decay factor $\gamma$ is set to 0.98. During rollout, the maximum generation length is 6,144 tokens, and the maximum number of interaction turns is 15. All training experiments are conducted on 8 NVIDIA H200 GPUs.

\subsection{Student Agent}
The student simulator is also implemented based on Qwen3-30B-A3B-Instruct-2507 \citep{yang2025qwen3}. Benefiting from the proposed cognition-decision decoupling mechanism, the simulator can effectively capture diverse student behaviors, eliminating the need for dedicated fine-tuning of the student model.

\subsection{Prompts}
\label{sec_appendix:prompt}
The complete prompt settings used in our method, including those for the Student Agent, Tutor Agent, and the Judge Agent can be found in the code. All external datasets and models are used only for research purposes and remain subject to their original licenses and terms of use. The released code and prompts are intended to support reproducibility and further research on LLM-based tutoring agents.

\subsection{Human Preference Annotation}
\label{sec_appendix:human_annotation}
We conduct two human preference evaluations. First, we randomly sample 300 dialogue pairs to assess agreement between reward-model predictions and human pedagogical judgments. Each pair contains two tutoring dialogues generated for the same problem and simulated-student condition. For diagnostic analysis, pairs are grouped into low-, medium-, and high-margin categories based on score differences assigned by Gemini-3.1-Pro. Second, teachers compare complete PEARL dialogues with those generated by GPT-5.1, Gemini-3-Flash, InnoSpark-plus-72B, MuduoLLM-14B, and Qwen3-30B-A3B under matched conditions. Model identities are hidden, A/B order is randomized, and outcomes are recorded as win, tie, or loss. Two experienced teachers independently annotate each pair using the problem, reference solution, student profile, and anonymized dialogues. Disagreements are resolved through discussion.

\subsection{Compute and Reproducibility Artifacts}
The full-parameter training pipeline uses eight NVIDIA H200 GPUs. Under the reported settings, tutor, student, and reward-model inference can each run on a single A100 GPU. We plan to release the 193K cold-start dataset, 803K reward-model annotations, the 300-pair preference set, complete prompts, and data-construction and training scripts.

\section{Case Studies}
\begin{figure*}[t]
	\centering
	\includegraphics[width=0.75\linewidth]{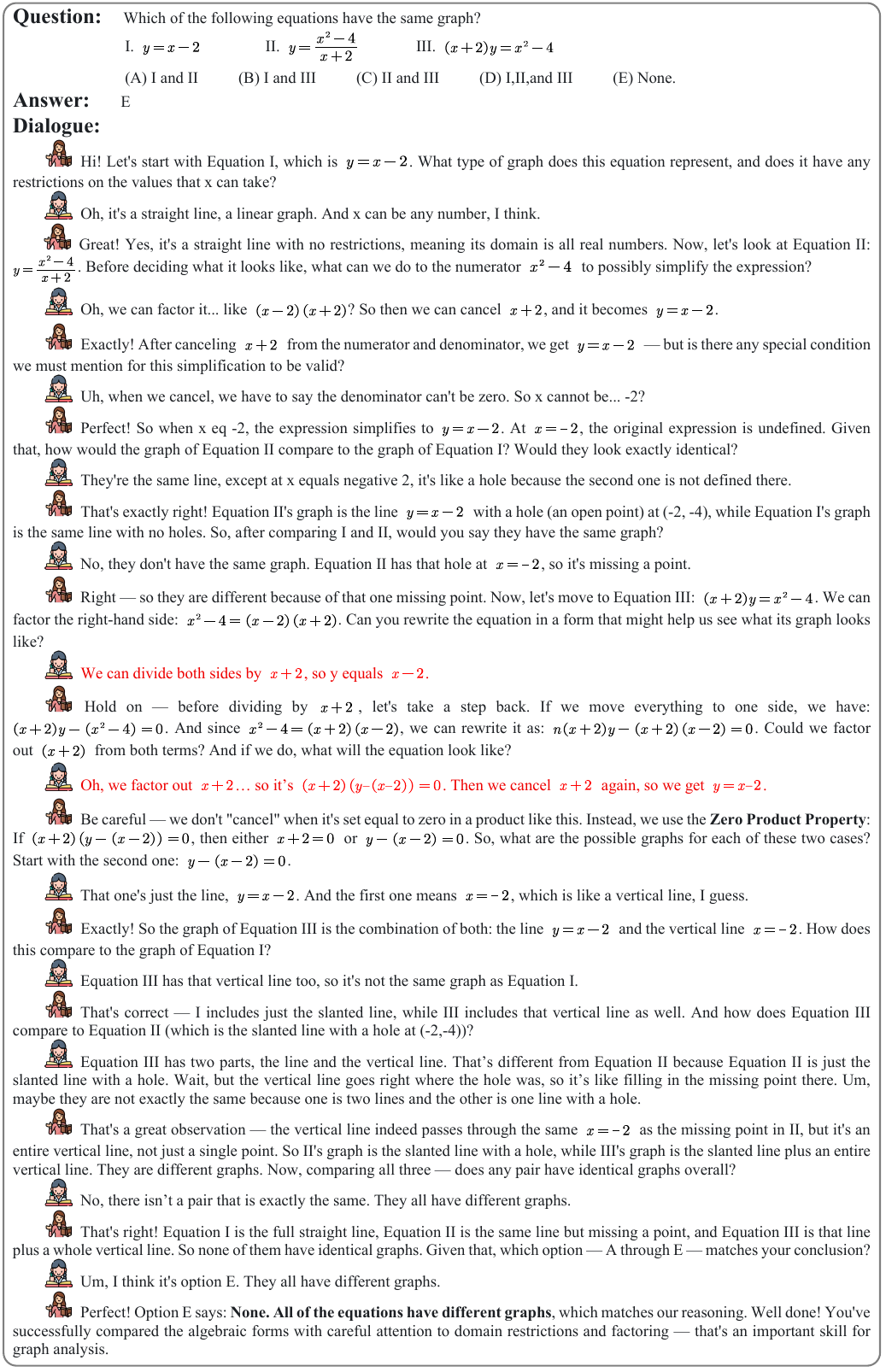}
	\caption{Case 1. The Student Agent exhibits plausible algebraic misconceptions by overlooking domain restrictions after canceling $x+2$, while the optimized tutor identifies the issue and guides the student to distinguish a complete line, a line with a hole, and a line together with a vertical line.}
	\label{fig:case_study_1}
\end{figure*}

\begin{figure*}[t]
	\centering
	\includegraphics[width=0.85\linewidth]{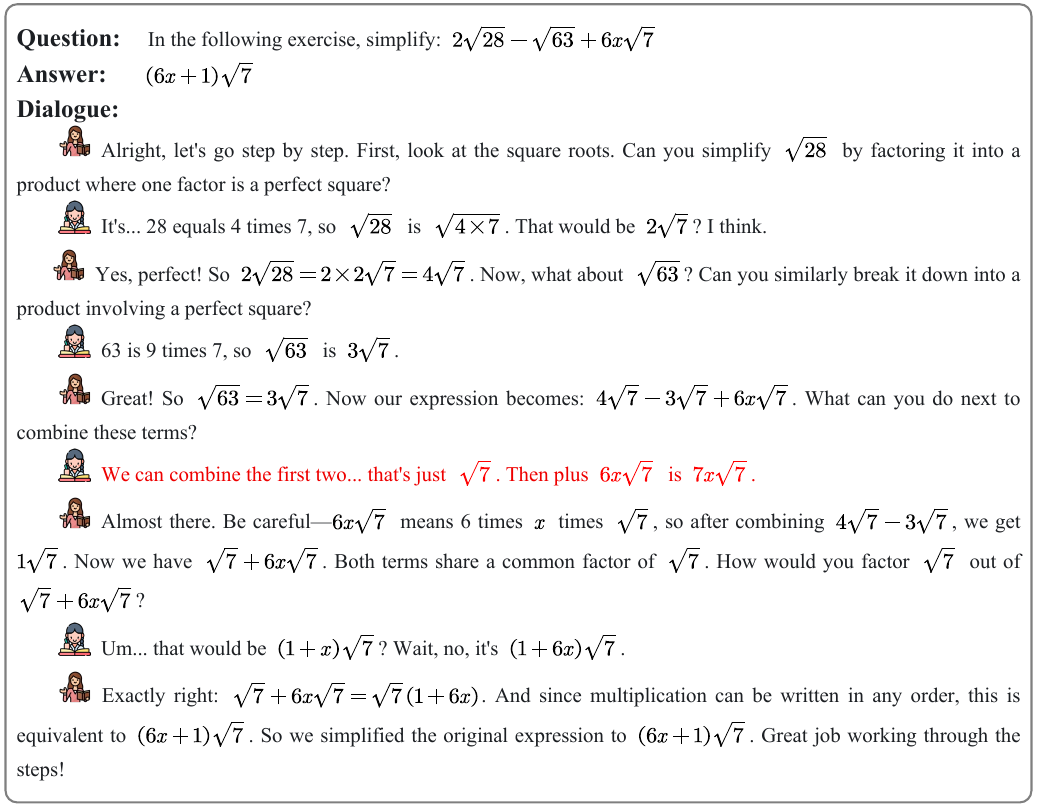}
	\caption{Case 2. The Student Agent plausibly miscombines $\sqrt{7}$ and $6x\sqrt{7}$ as $7x\sqrt{7}$, while the optimized tutor detects the error and guides the student to factor out the common radical.}
	\label{fig:case_study_2}
\end{figure*}

This section presents two case studies illustrating how PEARL guides students through Socratic tutoring, as shown in Figures~\ref{fig:case_study_1} and~\ref{fig:case_study_2}. In the two examples, the simulated Student Agent exhibits different misconceptions, while the tutor identifies the errors, provides timely feedback, and guides the student to reason through the problems step by step.

\section{Ethical Considerations}
This work studies LLM-based Socratic tutoring in simulated environments and does not use private student data. Human expert annotations are used only for preference evaluation and do not involve personally identifiable information. We inspect the generated tutoring data to remove potential personal identifiers or inappropriate content. Since LLM tutors may still produce incorrect or biased guidance, real-world deployment should require human oversight and appropriate safeguards.

\section{LLM Usage}
During manuscript preparation, we used large language models, including GPT-5.5 and Gemini-3.1-Pro, solely for language editing, such as improving grammar, readability, and clarity. The research ideas, experimental design, implementation, results, and scientific analyses presented in this work were developed and conducted by the authors.

\end{document}